\documentclass{article}

\usepackage{microtype}
\usepackage{graphicx}
\usepackage{subfigure}
\usepackage{booktabs} 
\usepackage{amsmath}
\usepackage{amssymb}
\usepackage{comment}
\usepackage{color}
\usepackage{xcolor}  
\usepackage{arydshln}
\usepackage{algorithm}
\usepackage{algorithmic}
\usepackage{paralist}
\usepackage{enumitem}
\usepackage{multirow}
\usepackage{dashrule}

\usepackage{hyperref}

\usepackage[accepted]{icml2023}

\usepackage{mathtools}
\usepackage{amsthm}

\usepackage[capitalize,noabbrev]{cleveref}

\theoremstyle{plain}

\theoremstyle{definition}

\theoremstyle{remark}

\usepackage[textsize=tiny]{todonotes}
\usepackage{makecell}

\icmltitlerunning{ESC: Exploration with Soft Commonsense Constraints for Zero-shot Object Navigation}

\begin{document}

\twocolumn[
\icmltitle{ESC: Exploration with Soft Commonsense Constraints \\ for Zero-shot Object Navigation}



\icmlsetsymbol{equal}{*}

\begin{icmlauthorlist}
\icmlauthor{Kaiwen Zhou}{yyy}
\icmlauthor{Kaizhi Zheng}{yyy}
\icmlauthor{Connor Pryor}{yyy}
\icmlauthor{Yilin Shen}{comp}
\icmlauthor{Hongxia Jin}{comp}
\icmlauthor{Lise Getoor}{yyy}
\icmlauthor{Xin Eric Wang}{yyy}
\end{icmlauthorlist}

\icmlaffiliation{yyy}{University of California, Santa Cruz}
\icmlaffiliation{comp}{Samsung Research America}

\icmlcorrespondingauthor{Xin Eric Wang}{xwang366@ucsc.edu}

\icmlkeywords{Machine Learning, ICML}

\vskip 0.3in
]



\printAffiliationsAndNotice{}  

\begin{abstract}
The ability to accurately locate and navigate to a specific object is a crucial capability for embodied agents that operate in the real world and interact with objects to complete tasks.
Such object navigation tasks usually require large-scale training in visual environments with labeled objects, which generalizes poorly to novel objects in unknown environments.  
In this work, we present a novel zero-shot object navigation method, Exploration with Soft Commonsense constraints (ESC), that transfers commonsense knowledge in pre-trained models to open-world object navigation without any navigation experience nor any other training on the visual environments.
First, ESC leverages a pre-trained vision and language model for open-world prompt-based grounding and a pre-trained commonsense language model for room and object reasoning. Then ESC converts commonsense knowledge into navigation actions by modeling it as soft logic predicates for efficient exploration.  
Extensive experiments on MP3D~\cite{Matterport3D}, HM3D~\cite{ramakrishnan2021hm3d}, and RoboTHOR~\cite{RoboTHOR} benchmarks show that our ESC method improves significantly over baselines, and achieves new state-of-the-art results for zero-shot object navigation (e.g., 288\% relative Success Rate improvement than CoW~\cite{gadre2022cow} on MP3D).~\footnote{\href{https://sites.google.com/ucsc.edu/escnav/home}{https://sites.google.com/ucsc.edu/escnav/home}}
\end{abstract}

\section{Introduction}

\begin{figure}[t]
  \centering
  \includegraphics[width=0.5\textwidth]{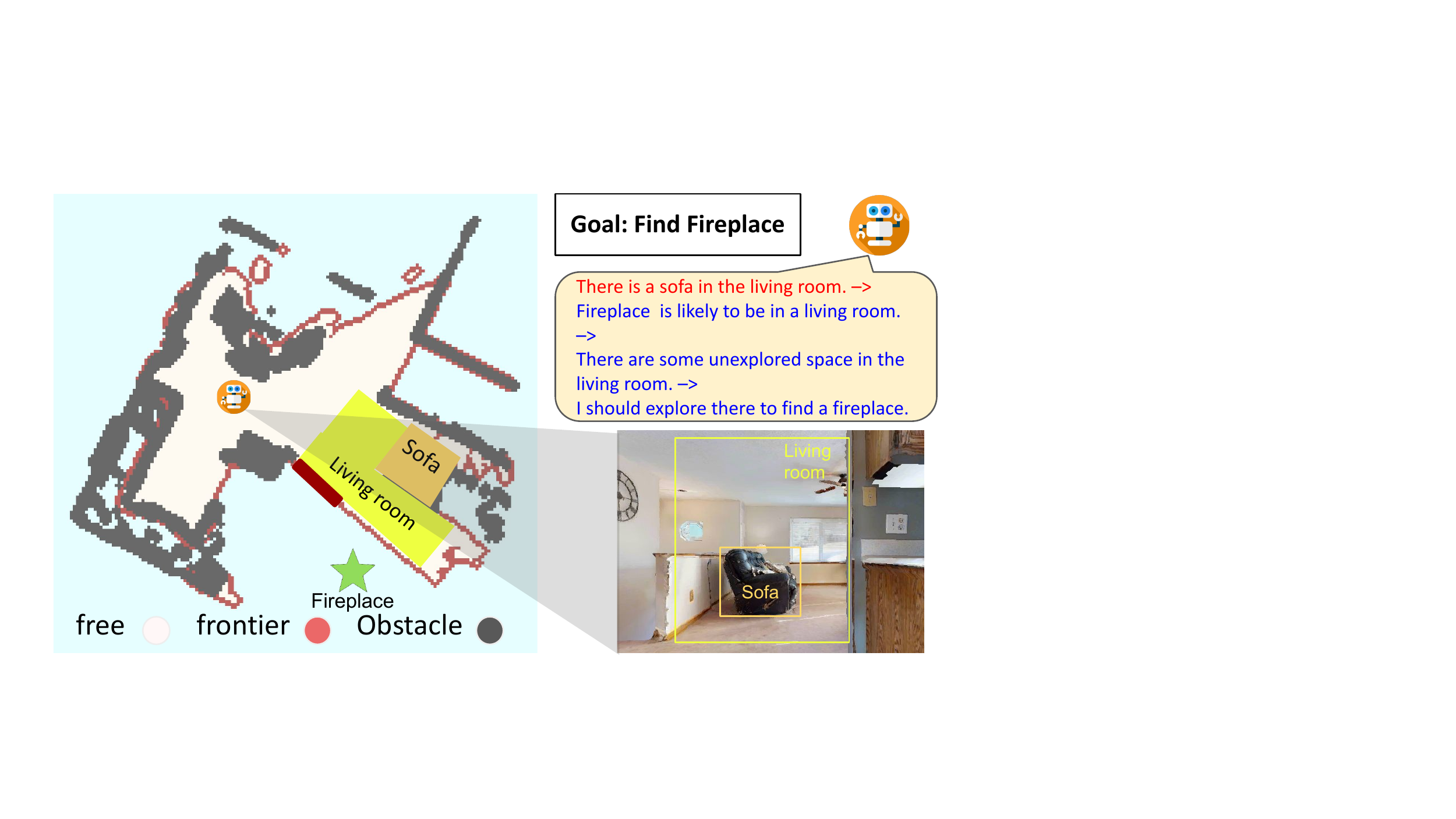}
     \vspace{-12pt}
   \caption{\textbf{Commonsense reasoning in object navigation.} 
   In object navigation, our agent first does a semantic understanding of the current scene (\textcolor{red}{red text} in the figure) and then performs commonsense reasoning (\textcolor{blue}{blue text} in the figure). The agent reasons that a fireplace is likely to be in a living room, so it decides to explore the unobserved part of the living room (the frontier adjacent to the observed part of the living room). }
   \label{fig:overview}
    \vspace{-10pt}
\end{figure}

Object navigation (ObjNav) is a task in which an embodied agent must navigate to a specific goal object within an unknown environment~\cite{objnav}. This task is fundamental to other navigation-based embodied tasks because navigating to a goal object is the preliminary for the agent to interact with it. 
While current state-of-the-art methods for object navigation achieve good results when trained on specific datasets with limited goal objects and similar environments, they often perform poorly when faced with novel objects or environments due to distribution shifts. Real-world situations often involve diverse objects and varied environments, making it difficult and costly to collect extensive, annotated trajectory data. As a result, generalized zero-shot object navigation, in which the navigation agent can adapt to novel objects and environments without additional training, is a crucial area of study.

Successfully navigating to a goal object requires two key abilities, (1) \textit{semantic scene understanding}, which involves identifying objects and rooms in the environment, and (2) \textit{commonsense reasoning}, which involves making logical inferences about the location of the goal object based on commonsense knowledge. 
For example, as in Fig.~\ref{fig:overview}, a fireplace is very likely in a living room, so the agent decides to explore the unseen area in the living room to find a fireplace.
However, current zero-shot object navigation methods have not yet effectively addressed this requirement and often lack commonsense reasoning abilities. 
Existing methods require training on other goal-oriented navigation tasks and environments~\cite{majumdar2022zson, zero_experience}, or use simple heuristics for exploration~\cite{gadre2022cow}.

Recent studies~\cite{he2021debertav3,clip,kojima2022large,li2021grounded} show that large pre-trained models have a strong generalization and reasoning ability for novel tasks under zero-shot scenarios.
Building upon this success, in this work, we propose a zero-shot object navigation framework, named Exploration with Soft Commonsense constraints (ESC), that leverages these pre-trained models and can seamlessly generalize to unseen environments and novel object types. 
As shown in Fig.~\ref{fig:overview}, we first use a prompt-based vision-and-language grounding model GLIP~\cite{li2021grounded} for open-world object grounding and scene understanding, which can infer the object and room information of current agent views. Benefiting from large-scale image-text pre-training, GLIP can easily generalize to new objects via prompting. Then, we utilize a pre-trained commonsense reasoning language model that takes the room and object information as context to infer the correspondence between rooms and objects.

However, there still remains a gap in converting the commonsense knowledge inferred from large language models (LLMs) into executable actions. 
In addition, the relationship between entities is usually uncertain, e.g., the book has a high probability in the living room, but it is not deterministic.
To address these challenges, our ESC method models ``soft'' commonsense constraints using Probabilistic Soft Logic (PSL)~\cite{psl}, a declarative templating language that defines a special class of Markov random fields with first-order logical rules.
Those soft commonsense constraints are then incorporated into a classic exploration method, frontier-based exploration (FBE), to determine which frontier to explore next in a zero-shot manner. 
Unlike previous methods that rely on implicit training of commonsense using neural networks~\cite{yang2019visual, chaplot2020object}, our method explicitly uses soft logic predicates to represent knowledge in a continuous value space, which is then assigned to each frontier, enabling more effective exploration. 

We demonstrate the effectiveness of our framework on three object goal navigation benchmarks, MP3D~\cite{Matterport3D}, HM3D~\cite{ramakrishnan2021hm3d}, and RoboTHOR~\cite{RoboTHOR}, with different house sizes, styles, texture features, and object types. 
Compared with CoW~\cite{gadre2022cow} that has the same setting as ours, our method achieves around 285\% relative improvement in success rate weighted by length (SPL) and success rate (SR) on MP3D and 35\% relative improvement in SPL and SR on RoboTHOR. Compared with ZSON~\cite{majumdar2022zson} that requires training on the HM3D dataset, our method outperforms it by 196\% relative SPL on MP3D and 85\% relative SPL on HM3D. Note that on the MP3D dataset, our zero-shot method is comparable with previous state-of-the-art supervised methods and achieves the best SPL.

In summary, our contributions are threefold:
\begin{itemize}
  \item We propose the Exploration with Soft Commonsense constraints (ESC) method for zero-shot object navigation, which leverages pre-trained vision and language models for open-world scene understanding and object-level and room-level commonsense reasoning. 
  \item Our ESC approach models soft commonsense constraints and seamlessly converts them into navigation actions using Frontier-based Exploration and Probabilistic Soft Logic, which is training-free.
  \item We achieve state-of-the-art results on zero-shot object goal navigation and outperform baseline methods by a large margin across three object navigation datasets and benchmarks. 
\end{itemize}


\section{Problem Definition} \label{sec:zson def}
In the conventional task of \textit{object navigation}, an agent is randomly placed within an unseen environment $E$ with a specified object category $G$ as a goal to find (e.g., chair, fireplace, or cabinet). The agent's objective is to navigate to any instance of the object that belongs to the aforementioned category.
At each time step $t$, the agent is presented with an observation $\mathcal{O}$, which consists of an egocentric RGB-D image $I_{t}$ and in some benchmarks, pose readings $P_{t}$. The agent needs to select an action $a$ from the action space $\mathcal{A}$, which includes a `STOP' action to terminate the navigation process. The navigation is considered successful if the agent stops within $d_{s}$ meters of the object and the object is visible without further moving.

In contrast to supervised object navigation, which trains the agent on the objects and environments it will navigate, this work focuses on \emph{zero-shot object navigation}: given a new set of environments $\{E_{new}\}$ and a new set of goal objects $\{o_{new}\}$ that the agent has not seen before, the agent is required to perform object goal navigation in $\{E_{new}\}$ for $\{o_{new}\}$ without training on object goal related labels. Furthermore, we target an even more challenging zero-shot scenario---the agent performs zero-shot object navigation without training on any navigation data.

\begin{figure*}[t]
  \centering
  \includegraphics[width=0.85\textwidth]{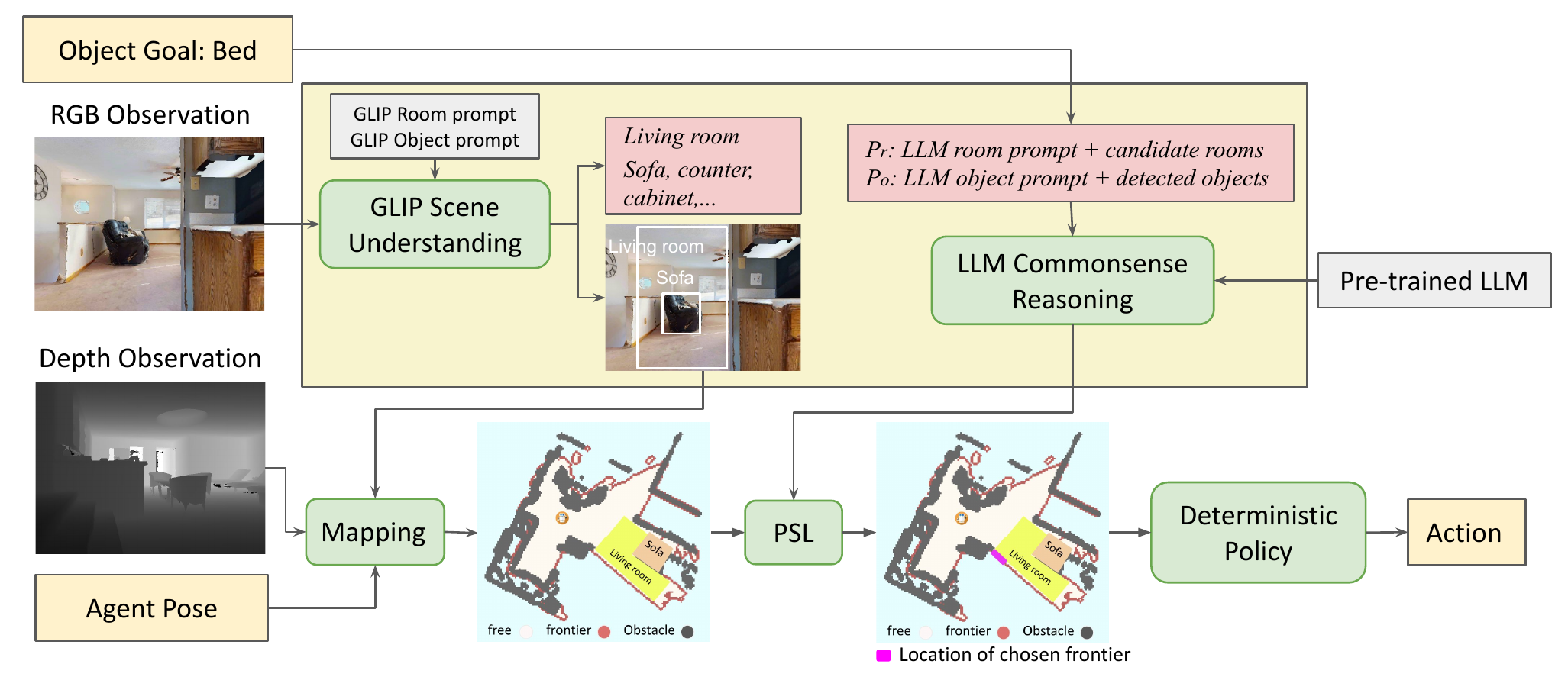}
    \vspace{-1ex}
   \caption{\textbf{The ESC framework.} During navigation, the agent performs scene understanding based on RGB observations and prompts. Meanwhile, the Mapping module constructs a semantic map containing room, object, and frontier information. Conditioned on the goal object and semantic scene information, the agent will then perform commonsense reasoning via a LLM to infer the probable location of the goal object, and select a frontier to explore using PSL. 
   }
   \label{fig:framework}
     \vspace{-10pt}
\end{figure*}

\section{Our ESC Approach} \label{sec:method}

In this section, we outline our Exploration with Soft Commonsense constraints (ESC) framework for zero-shot object navigation. As in Fig.~\ref{fig:framework}, the ESC framework first converts the input image into a semantic understanding of the scene and projects it to a semantic map (Sec.~\ref{sec:glip}). Then it leverages large language models to perform commonsense reasoning for the spatial relations between the goal object and common objects and rooms (Sec.~\ref{sec:llm}). Lastly, it combines frontier-based exploration with semantic scene understanding and commonsense reasoning via PSL (Sec.~\ref{sec:commonsense nav}). 

\subsection{Open-World Semantic Scene Understanding} \label{sec:glip}
\textbf{Prompt-Based Scene Grounding} 
To leverage large language models for navigation inference, we need to transform the input RGB images into semantic context in language form. To achieve this, we leverage a pre-trained grounded language-image model GLIP~\cite{li2021grounded} using a text prompt.
Unlike traditional object detection models such as Mask-RCNN~\cite{mrcnn}, which is limited to fixed classes, GLIP formulates the detection task as a grounding problem by aligning the proposed image region with phrases in the text prompt and predicting the score of region-text alignment. 
Benefiting from large-scale image-text pretraining, GLIP can detect common indoor concepts (\textit{e.g.} object, room) in an open-world setting.
And it is easy to generalize to different environments and object goals to perform open-world object navigation.

We first define a set of common indoor objects $\{o_c\}$~\footnote{The full list is shown in Appendix~\ref{sec:glip appendix}.}, then we take the union of $\{o_c\}$ and all the possible goal objects $\{o_g\}$ to generate an object prompt for object grounding. The object prompt $P_{o}$ will be the object names in $\{o_c\}\cup \{o_g\}$ joined by `. '. 
For example, if $\{o_c\}=\{cabinet, table\}$ and $\{o_g\} = \{chair, table\}$, then the object prompt $P_{o}$ will be `cabinet. chair. table.'. 

However, object information in the current scene is a relatively low-level scene context. When humans search for a goal in an unseen environment, they will usually consider higher-level contexts (e.g., which room is likely to contain the goal?).
Therefore, we define a set of common rooms $\{r_c\}$ in indoor environments for room prompt $P_{r}$ to detect room information.

By inputting these prompts and an egocentric image into the GLIP model as in Fig.~\ref{fig:framework}, we can get the detected objects $o_{t,i}$, rooms $r_{t,i}$ and bounding boxes from the current scene:
\begin{gather}
    \{o_{t,i}, b^{o}_{t,i}\}=\mathrm{GLIP}(I_{t}, P_{o}) \label{eq1}\\
    \{r_{t,i}, b^{r}_{t,i}\}=\mathrm{GLIP}(I_{t}, P_{r})
\end{gather}
where $b^{o}_{t,i}$ and $b^{r}_{t,i}$ are the bounding boxes of the objects and rooms. Notice that these prompts can be easily extended to generalize to new test data to perform open-world semantic scene understanding. 

\textbf{Semantic Map Construction} 
Based on the depth input $D_{t}$, agent location, and camera parameters, we can transform the pixels in a 2D image into 3D space, which is stored in a 3D voxel, where transformed pixels close to the floor are considered free space. Then we project the 3D voxel along the height dimension and obtain a 2D navigation map as shown in Fig.~\ref{fig:framework}, which will be maintained during navigation. Through the navigation map, we can obtain the frontiers in the current map as explained in Sec.~\ref{sec:fbe intro}. 

Furthermore, as shown in Fig.~\ref{fig:framework}, we can project the detected room and object location into a semantic map. For object detection, we take the center of a bounding box and project it to a 2D location. For room detection, we project all the pixels in a bounding box into a 2D map and record the projected locations as the corresponding room.

\subsection{Commonsense Reasoning for ObjNav via LLM} \label{sec:llm}
In an in-door environment, a goal object will appear in certain rooms and near certain objects more frequently, and this kind of common sense is helpful for the agent to search for a goal object. Therefore, after detecting the room and object information in the current scene, we can leverage pre-trained large language models to perform commonsense reasoning conditioned on the goal object and semantic scene information via text prompt. 

Specifically, for object-level and room-level inference, the large language models can reason on whether a goal object $G$ is likely to be near each object $o_{i}$ in the object prompt $P_{o}$ and whether it is likely to be in each room $r_{i}$ in the room prompt $P_{r}$. The prediction output of the large language models will be the real-value scores $S(G|o_{i}), S(G|r_{i}) \in [0,1]$ of each (goal, object) pair and (goal, room) pair. How to get the scores from language models and the text prompt for the language model can vary between different LLMs. 
We mainly use Deberta v3~\cite{he2021debertav3} in our method for its effectiveness and accessibility. Details about the LLMs can be found in Appendix~\ref{sec:llm appendix}. 

\subsection{Commonsense Guided Exploration} \label{sec:commonsense nav}
\subsubsection{Frontier-based Exploration}
\label{sec:fbe intro}
In object goal navigation, exploring the environment efficiently is very important to find the target object, as the object is often not seen in the agent's initial location. \citet{gadre2022cow} use a heuristic exploration method, Frontier-based Exploration (FBE), to explore the environment, which shows superior performance compared with learning-based exploration methods. 
As shown in Fig.~\ref{fig:overview}, a frontier in a map is defined as the border between the free area and the unseen area. Free area is defined as the area that the agent has seen and is not occupied by obstacles. One common strategy for frontier selection is to choose the closest frontier (with a distance threshold $d_{f}$) as the next subgoal~\cite{gadre2022cow}.
However, choosing the closest frontier as a subgoal to explore may not be optimal in semantic-rich environments and may be against commonsense. 
For example, the agent might check the frontiers behind the couch in a living room to search for a bed. 

Therefore, we propose to introduce commonsense knowledge in LLMs into frontier-based exploration. Our goal is to make the frontier selection decision based on not only the distances $d_{i}$ from the agent but also object $o^{t}$ and room $r^{t}$ information around the frontiers:
\begin{equation}
    P(F) = P(F|d_{i}, o^{t}, r^{t})
\end{equation}
Intuitively, we are more likely to choose a frontier close to an object near which the goal object is likely to appear, or a frontier in a room in which the goal object should be. 
Notice that this kind of rule represents a concept that is not always correct, as there may be multiple frontiers satisfying potentially many rules, and the conditions within these rules are continuously valued. Therefore, we need a system that can express these soft rules and logic well, i.e., Probabilistic Soft Logic (PSL).

\subsubsection{Soft Commonsense Constraints} 
Now we describe how we combine commonsense reasoning with frontier-based exploration mentioned above via PSL and enable frontier selection with soft commonsense constraints.
Probabilistic Soft Logic (PSL)~\cite{psl} is a probabilistic programming language defining hinge-loss Markov random fields (HL-MRF) using a syntax based on first-order logic. Specifically, PSL models dependencies between relations and attributes of entities in a domain, defined as \textit{atoms}, which are encoded with weighted first-order logical clauses and linear arithmetic inequalities referred to as \textit{rules}. 
We define four following rules for object navigation. 

\textbf{a. Object reasoning} We first consider object-level reasoning. 
To encourage the agent to explore those frontiers near some objects that are likely to appear around the goal object, we have the rule:
\begin{equation} \label{eq: obj reason 1}
\begin{aligned}
    \mathit{w}: \mathrm{IsCooccur(Goal, Object)} \\
    \land \ \mathrm{IsNearObj(Frontier, Object)} \\ 
    \longrightarrow \mathrm{ChooseFrontier(Frontier)}
\end{aligned}
\end{equation}
The parameter $\mathit{w}$ is the weight of the rule, quantifying its relative importance in the model. This rule includes three atoms: $\mathrm{IsCooccur(Goal, Object)}$, $\mathrm{IsNearObj(Frontier,}$ $\mathrm{Object)}$, and $\mathrm{ChooseFrontier(Frontier)}$. 
The value of $\mathrm{IsCooccur(Goal, Object)}$ is the co-occurrence score $S(G|o_{i})$ for (Goal, Object) pair predicted by the language models. The value of $\mathrm{IsNearObj(Frontier, Object)}$ is the confidence of the object prediction by GLIP if the object is within $d_{o}$ meters of the frontier according to the semantic map from Sec.~\ref{sec:glip}; otherwise, $\mathrm{IsNearObj(Frontier, Object)}=0$. Furthermore, to discourage the agent from going to those frontiers near some objects that are unlikely to be around the goal object, we have a corresponding negative rule:
\begin{equation} \label{eq: obj reason 2}
\begin{aligned}
    \mathit{w}: \mathrm{!IsCooccur(Goal, Object)} \\
    \land \ \mathrm{IsNearObj(Frontier, Object)} \\ 
    \longrightarrow \mathrm{!ChooseFrontier(Frontier)}
\end{aligned}
\end{equation}

\textbf{b. Room reasoning} Similar to object reasoning, we encourage the agent to explore the frontiers near or in a room where the goal object is likely to appear, and discourage it from exploring the frontiers near or in a room where the goal object is unlikely to appear. Thus, we have two rules for room reasoning similar to Eq.~(\ref{eq: obj reason 1}) and Eq.~(\ref{eq: obj reason 2}) where `Object' is substituted with `Room'.

\textbf{c. Distance constraint} The vanilla frontier-based exploration method~\cite{gadre2022cow} chooses the frontier with the shortest distance from the agent, which encourages the agent to continue exploring one area until there is nothing to explore. We also add a shortest-distance rule to encourage the agent to explore nearby frontiers:
\begin{equation}
\begin{aligned} 
    \mathit{w}: \mathrm{ShortDist(Frontier)} \\
    \longrightarrow \mathrm{ChooseFrontier(Frontier)}
\end{aligned} \label{eq: shortest dist}
\end{equation} 

\textbf{d. Sum constraint} We adapt a PSL hard constraint to limit the sum of the scores of choosing all the frontiers to one:
\begin{equation}
\begin{aligned}
 \mathrm{ChooseFrontier(+Frontier)} = 1
\end{aligned}
\end{equation}
This constraint prevents the degenerated solution where all the target variables are equal to one and encourages the frontiers to compete with each other. 

\textbf{PSL Inference}
During PSL inference, atoms will be instantiated with data and referred to as ground atoms. Taking equation~\ref{eq: obj reason 1} as an example. Ground atoms are mapped to either an observed variable $X$, like $\mathrm{IsCooccur}$ and $\mathrm{IsNearObj}$, or a target variable $Y$, like $\mathrm{ChooseFrontier}$. Then, valid combinations of ground atoms substituted in the rules create ground rules. Each ground rule creates one or more hinge-loss potentials defined over logical rules, which are relaxed using Łukasiewicz continuous valued logical semantics:
\begin{equation}
    \phi (Y,X) = [max(0,l(Y,X))]^{p}
\end{equation}
where $l$ is a linear penalty function\footnote{Details of the penalty function can be found in Appendix~\ref{sec:psl appendix}.} defined by PSL. $\phi (Y,X)$ represents the \textit{distance to satisfaction} of this ground rule. The values of $X,Y$ are in the range $[0,1]$, and $p\in \{1,2\}$ optionally squares the potentials. 

Given all the observed variables $X$ and target variables $Y$, PSL defines an HL-MRF over the target variables:
\begin{equation}
    P(Y|X) = \frac{1}{Z(Y)} exp(-\sum^{m}_{i=1}w_{i}\phi_{i}(Y,X))
\end{equation}
\begin{equation}
    Z(Y) = \int_{Y} exp(-\sum^{m}_{i=1}w_{i}\phi_{i}(Y,X))
\end{equation}
where $m$ denotes the number of potential functions, $\phi_{i}$ is the $i^{th}$ potential function, $w_{i}$ is the weight of the template rule for $\phi_{i}$. 

Therefore, the optimization for the distribution can be converted to a convex optimization problem:
\begin{align} \label{eq: psl objective}
    Y^{*} &= \underset{Y}{argmin} \sum^{m}_{i=1}w_{i}\phi (Y,X)
\end{align}

\textbf{One-hot constraint PSL solver} Normally, a PSL program can use convex optimization algorithms such as ADMM~\cite{admm} to find a solution. However, since the final choice of the agent is only one frontier, we further limit the solution space to one hot encoding space, where each one-hot encoding represents selecting one frontier. Our one-hot constraint solver is performed by calculating the violation of constraints for each one-hot encoding. The encoding with the lowest loss, representing the frontier with the lowest value of violated constraints, will be chosen. This approach can help us save the iteration time of optimization compared with convex optimization algorithms. 


\textbf{Navigation Policy} The agent adapts a simple navigation policy with commonsense reasoning. It will choose a new frontier based on PSL inference after it reaches a frontier. After the agent detect the goal object, it will directly navigate to it. The agent is equipped with a deterministic policy as in Fig.~\ref{fig:framework} to help it navigate to the goal or a frontier. The full navigation algorithm is in Appendix~\ref{sec: nav alg appendix}.

\begin{table*}[t]
\centering
\setlength{\abovecaptionskip}{8pt}
\setlength{\belowcaptionskip}{8pt}
\caption{\textbf{Zero-shot object navigation results} on MP3D~\cite{Matterport3D}, HM3D~\cite{ramakrishnan2021hm3d}, and RoboTHOR~\cite{RoboTHOR} benchmarks. Notice that our method and CoW~\cite{gadre2022cow} are the only two zero-shot methods with no navigation training experience. Our method significantly outperforms previous zero-shot methods. * The training environment of ProcTHOR is similar to RoboTHOR using the same simulator. 
}
\vskip 0.1in
\resizebox{0.95\textwidth}{!}{
\begin{tabular}{lccccccccc}
\toprule
\multirow{2}*{\textbf{Model}} & \multirow{2}*{\textbf{Supervised}} & \multirow{2}*{\textbf{Trained on environment}} & \multirow{2}*{\textbf{Navigation training}} &
    \multicolumn{2}{c}{\textbf{MP3D}} & \multicolumn{2}{c}{\textbf{HM3D}} & \multicolumn{2}{c}{\textbf{RoboTHOR}}\\
    \cmidrule(lr){5-6}\cmidrule(lr){7-8} \cmidrule(lr){9-10}
    & & & & SPL$\uparrow$ & SR$\uparrow$ & SPL$\uparrow$ & SR$\uparrow$ &  SPL$\uparrow$ & SR$\uparrow$  \\
    \midrule
$\mathrm{PONI}$~\cite{ramakrishnan2022poni} & Yes &Yes & No &12.1 & 31.8 & - & - & - & -\\
$\mathrm{ProcTHOR}$~\cite{procthor}  & Yes &Yes &Yes& - & - & 31.8 & 54.4 & 28.8 & 65.2\\ 
\midrule
$\mathrm{ZSON}$~\cite{majumdar2022zson} &No &Yes &Yes & 4.8 & 15.3 & 12.6 & 25.5 & - & -  \\
$\mathrm{ProcTHOR-ZS}$~\cite{procthor}  &No &No* &Yes & - & - & 7.7 & 13.2 & \textbf{23.7} & \textbf{55.0} \\
\cdashline{1-10}
$\mathrm{CoW}$~\cite{gadre2022cow}  &No &No &No & 3.7 & 7.4 & - & - & 16.9 & 26.7 \\
ESC (Ours) &No &No &No & \textbf{14.2} & \textbf{28.7} &\textbf{22.3}  & \textbf{39.2} & 22.2 &38.1\\
\bottomrule
\end{tabular}
}
\label{tab:main results}
  \vspace{-2ex}
\end{table*}

\section{Experimental Setup} \label{sec:experiments}
\subsection{Benchmarks and Metrics}
\vspace{0.2ex}
\noindent\textbf{MP3D}~\cite{Matterport3D} is used in Habitat ObjectNav challenges, containing 2195 validation episodes on 11 validation environments with 21 goal object categories. 

\vspace{0.2ex}
\noindent\textbf{HM3D}~\cite{ramakrishnan2021hm3d} is used in Habitat 2022 ObjectNav challenge, containing 2000 validation episodes on 20 validation environments with 6 goal object categories. 

\vspace{0.2ex}
\noindent\textbf{RoboTHOR}~\cite{RoboTHOR} is used in RoboTHOR 2020, 2021 ObjectNav challenge, containing 1800 validation episodes on 15 validation environments with 12 goal object categories. Different from HM3D and MP3D, the goal objects in RoboTHOR are mainly small objects. 

\vspace{0.2ex}
\noindent\textbf{Metrics~}
On all three benchmarks, the number of maximum navigation steps is 500. We report and compare Success Rate (SR) and Success Rate weighted by inverse path Length (SPL)~\cite{spl}, of which SPL is the primary metric used in the Habitat and RoboTHOR challenges. In ablation studies, we also report SoftSPL~\cite{softspl}, which reflects the navigation progress made by the agent considering navigation efficiency.

\vspace{0.2ex}
\noindent\textbf{Agent Configurations~}
The agent has a height of 0.88m, with a radius of 0.18m. The agent receives $640\times 480$ RGB-D egocentric views from a camera with $79^{\circ}$ HFoV placed 0.88m from the ground. All the agents have action space of $\mathcal{A}=\{$MoveForward, RotateRight, RotateLeft, LookUp, LookDown, Stop$\}$. The moving step is 0.25m, and each rotation turns the agent by $30^{\circ}$. In MP3D and HM3D datasets, the agent will receive its GPS location at each step. 

\subsection{Baselines}
We compare our ESC method with the following two state-of-the-art (SOTA) methods for zero-shot object navigation: 
\begin{itemize}
    \item \textbf{ZSON}~\cite{majumdar2022zson} uses a CLIP encoder to project the object and image goal into a same embedding space and feed the object goal embedding into an image goal navigation~\cite{image-goal} network.
  \item \textbf{CLIP on Wheels (CoW)}~\cite{gadre2022cow} use gradient-based visualization technique (Grad-CAM~\cite{gradcam}) on CLIP to localize goal object in egocentric view, and a frontier-based exploration technique~\cite{fbe} for zero-shot object goal navigation. 
\end{itemize}
In addition, we also compare our method with the following supervised methods:
\begin{itemize}
  \item \textbf{PONI}~\cite{ramakrishnan2022poni} proposes a modular method that predicts goal object potential and explorable area potential to select a temporary goal from the semantic map, achieving SOTA results on MP3D.
  \item \textbf{ProcTHOR}~\cite{procthor} synthesizes 10k indoor environments and performs large-scale ObjectNav training on those environments. Then it fine-tunes the agent on each specific dataset, achieving SOTA results on HM3D and RoboTHOR. We also compare with its zero-shot version in Table~\ref{tab:main results}. 
\end{itemize}

\subsection{Implementation Details}
There are several hyper-parameters in the ESC method. For the distance threshold $d_f$ for selecting the closest frontier to explore, we use $d_f = 1.6 m$ in all the experiments. For the threshold $d_o$ determining whether a frontier is near an object, we fix $d_o = 1.6 m$. For the threshold $d_r$ determining whether a frontier is in a room, we fix $d_r = 0.6 m$.  
We applied a weight of 1.0 for all PSL rules when only one of commonsense reasoning (object or room) was utilized. Moreover, we double the weight for the shortest distance rule in Eq.~\ref{eq: shortest dist} to 2.0 when both levels of commonsense reasoning are employed.

\section{Results and Analysis}
\label{sec:result analysis}

\begin{figure}[t]
  \centering
         \includegraphics[width=0.45\textwidth]{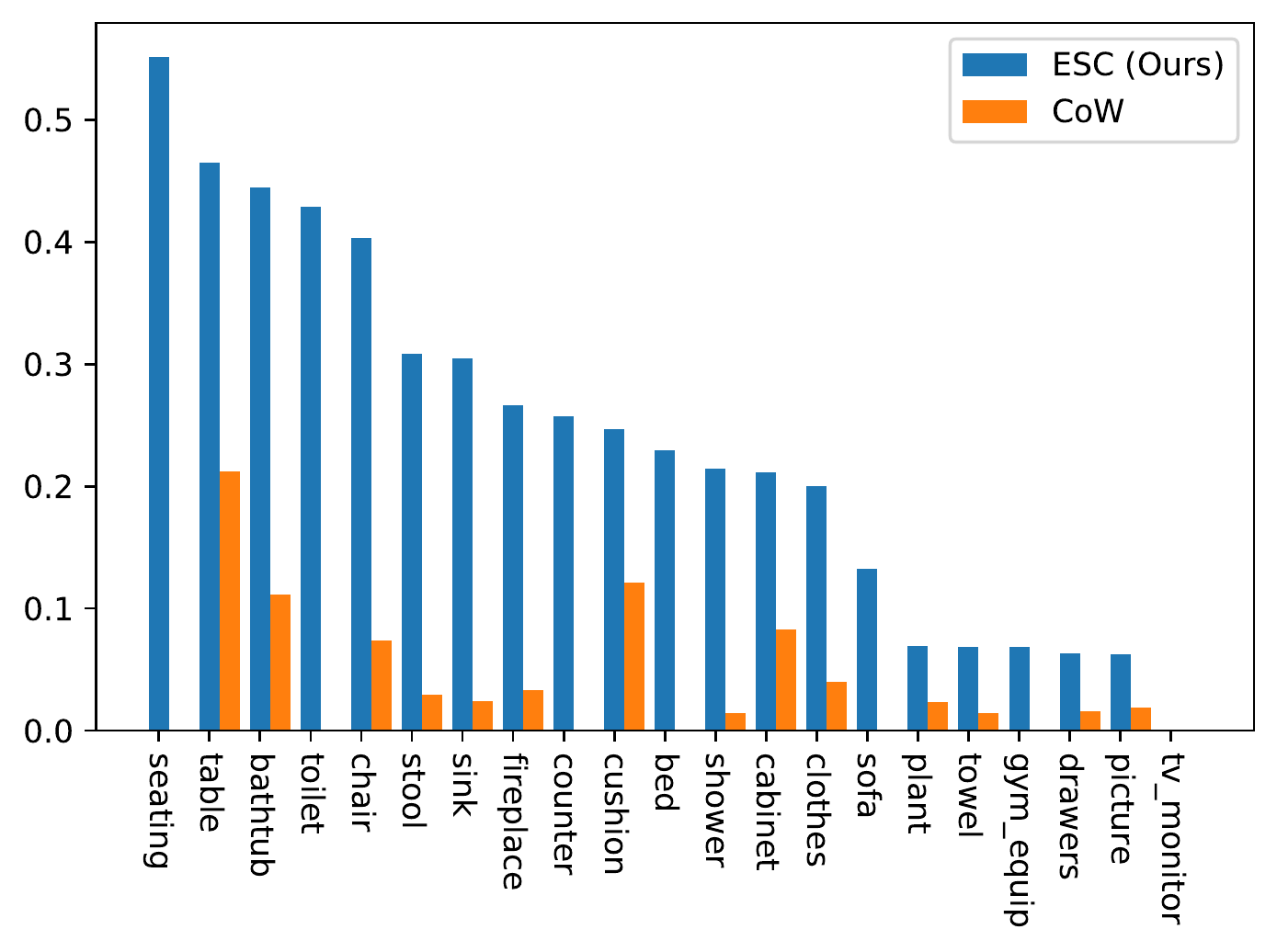}
    \vspace{-2ex}
  \caption{
  Comparison of the success rate of each goal category on MP3D between CoW~\cite{gadre2022cow} and ESC.}
  \label{fig:result by cat}
  \vspace{-15pt}
\end{figure}

\subsection{Result Comparison with SOTA Methods}
We compare the performance of our method with other zero-shot and supervised state-of-the-art (SOTA) methods in Table~\ref{tab:main results}. 
Our ESC method significantly outperforms previous zero-shot methods on both SR and SPL metrics, \textit{e.g.}, with 284\% SPL and 288\% SR relative improvements on MP3D over CoW that has the same evaluation setting with our method. 
Compared with ZSON~\cite{majumdar2022zson}, which trains the agent on HM3D datasets for the image-goal navigation task at a large scale, our method still outperforms it by a large margin. 
The zero-shot version of the ProcTHOR model is trained on ProcTHOR data and adapts to other datasets directly. Since the texture, style, and layout of the pre-training data of ProcTHOR are similar to RoboTHOR with the same simulator, the ProcTHOR agent achieves a much higher success rate on RoboTHOR dataset. However, ProcTHOR suffers from a severe performance drop on HM3D dataset, which is reconstructed from real-world architectures with photo-realistic images and on the Habitat simulator. ESC outperforms ProcTHOR on average on SR and SPL on these two datasets.



\begin{table*}[t]
\centering
\caption{Comparison between different detection models and different levels of commonsense reasoning on three datasets.}
\vskip 0.1in
\label{tab:ablation reasoning}
\resizebox{0.75\textwidth}{!}{
    \begin{tabular}{lcccccccc}
    \toprule
         & \multicolumn{3}{c}{\textbf{MP3D}} & \multicolumn{3}{c}{\textbf{HM3D}} & \multicolumn{2}{c}{\textbf{RoboTHOR}} \\
         \cmidrule(lr){2-4}  \cmidrule(lr){5-7} \cmidrule(lr){8-9} Commonsense
         Reasoning & SPL & SR & SoftSPL & SPL & SR & SoftSPL & SPL & SR\\
         \midrule
        CLIP w/o Commonsense ($\mathrm{CoW}$) & 3.7 & 7.4 & - & - & - & - & 16.9 & 26.7 \\
        GLIP w/o Commonsense ($\mathrm{GoW}$) & 12.4 & 25.6 & 22.8 & 18.8 & 33.1 & 27.0 & 21.6 & 36.1 \\
        ESC w/o Room  & 13.8  & 28.3  & 23.7 & 21.8  &\textbf{39.3}  & 30.6 &  21.4  & 35.6 \\
        ESC w/o Object  & 13.7  & 27.8  &\textbf{24.1} & \textbf{22.3}  & \textbf{39.3}  & 30.2 & \textbf{23.1} & \textbf{39.3} \\
        ESC & \textbf{14.2} &  \textbf{28.7} & 23.8 & \textbf{22.3} & 39.2 & \textbf{31.1} & 22.2 & 38.1 \\
        \bottomrule 
    \end{tabular}
    }
    \vspace{-12pt}
\end{table*}

\begin{table}[t]
\centering
\caption{Comparison between Deberta and ChatGPT on HM3D.}
\vskip 0.1in
\label{tab:ablation llm}
\resizebox{\columnwidth}{!}{
    \begin{tabular}{lcccccc}
    \toprule
         & \multicolumn{3}{c}{\textbf{Deberta}} & \multicolumn{3}{c}{\textbf{ChatGPT}}  \\
         \cmidrule(lr){2-4}  \cmidrule(lr){5-7} 
         Model & SPL & SR & SoftSPL & SPL & SR & SoftSPL \\
         \midrule
         Object & 21.8  & \textbf{39.3} & 30.6  & 22.1  & 38.8  & 30.8  \\
         Room & \textbf{22.3}  & \textbf{39.3}  & 30.2  & 21.6  & 38.0 & 29.7 \\
         Obj+Room & \textbf{22.3} & 39.2 & \textbf{31.1} & \textbf{22.4} & \textbf{39.0} & \textbf{31.1}\\
        \bottomrule 
    \end{tabular}
    }
    \vspace{-15pt}
\end{table}

Moreover, all three other zero-shot methods suffer from inconsistent performance when adapting to different datasets. This shows that their scene understanding and navigation policy are not generalized enough. Our ESC method, in contrast, performs consistently well on three datasets, demonstrating its strong generalizability. 

What is more, our ESC method significantly reduces the gap between zero-shot methods and supervised methods on HM3D and RoboTHOR datasets, and even outperforms the supervised THDA method~\cite{Treasure_Hunt} on MP3D, which shows the great potential of zero-shot methods on object goal navigation tasks. 

Fig.~\ref{fig:result by cat} illustrates a category-wise comparison between CoW and ESC on the MP3D dataset.
It is evident that ESC outperforms CoW consistently on all the goal object types. 
Note that CoW fails on some goal objects with a strong tendency to appear or not to appear in specific rooms or in proximity to particular objects (e.g., `toilet' and `bed'), while our agent performs much better in those cases, indicating the efficacy of commonsense reasoning.\footnote{There are only 7 `TV\_monitor' examples on MP3D, so its performance is not representative. ESC's SR for `TV\_monitor' on HM3D is 21.7\%, which has 281 `TV\_monitor' examples.}

\subsection{Ablation Study}
To demonstrate the efficacy of semantic scene understanding and commonsense reasoning, we design GLIP on Wheel (GoW). It uses a GLIP model for object detection and the vanilla fronter-based exploration method for exploration. As a replacement for commonsense reasoning in ESC, GoW always selects the closest frontier 1.6 meters away during exploration. Notice that GoW has the same navigation policy as ESC methods except the frontier selection policy. 

\textbf{Effect of semantic scene understanding and commonsense reasoning.} In Table~\ref{tab:ablation reasoning}, GoW surpasses CoW on all metrics on MP3D and RoboTHOR, which demonstrates the effectiveness of open-world object grounding of GLIP. 
Furthermore, ESC further outperforms GoW on all the datasets and metrics, showing the superiority of commonsense reasoning compared with pure heuristic exploration.

\textbf{Effect of different levels of commonsense reasoning.} In Table~\ref{tab:ablation reasoning}, we also compare the performance of different levels of reasoning on three datasets. We remove object/room level common sense for comparison. From the results, we observe that both room and object reasoning improves over GoW, and room reasoning usually brings larger improvement than object reasoning. Using both room and object reasoning provides better results on MP3D and HM3D datasets. 
Due to the more random placement of objects in the RoboTHOR dataset, exploration without object reasoning performs the best, and incorporating object reasoning hurts the performance slightly.

\textbf{Effect of different LLMs.} Table~\ref{tab:ablation llm} compares the performance of different LLMs for commonsense reasoning on HM3D.
Both LLMs significantly improve the performance over GoW. 
ChatGPT performs similarly to Deberta, except for using room-level commonsense even without specific commonsense training. We mainly use Deberta in our framework due to its accessibility. See Appendix~\ref{sec:llm appendix} for more implementation details.


\textbf{How commonsense reasoning helps exploration.}
In Table~\ref{tab:frontier dist}, we compare the exploration ability of GoW and ESC with different frontier selection strategies. GoW selects the closest frontier 1.6 meters away from the agent, while ESC selects the frontier based on the commonsense knowledge inferred from LLMs. 
First, we calculate the average distance of all the chosen frontiers to the closest target object of different methods. 
As shown in the first column of Table~\ref{tab:frontier dist}, the frontiers chosen by our ESC method are closer to the goal object on average, which indicates our method can perform better exploration consistently and help the agent get closer to the goal object.

Second, we demonstrate the error analysis of GoW and ESC models in Table~\ref{tab:frontier dist}. For failure navigation, we define three kinds of errors: 
\emph{Detection error} is defined as the goal appearing in the vision, but the agent didn't correctly detect it, or the goal never appears, but the agent thought it detected a goal. 
\emph{Planning error} is defined as the agent successfully detecting the target object but failing, or the agent never detecting the goal object and stuck within 1 meter for at least 400 steps, which indicates low-level navigation ability.
\emph{Exploration error} is an error that is not a planning or detection error, which means the agent never saw the goal object without stuck or false detection. The exploration error rate evaluates the ability to get close to the goal object. 

In Table~\ref{tab:frontier dist}, we observe that the exploration error of our ESC method has the most decrease compared with GoW. This validates that our ESC method helps the agent better explore the environment and get close to the object. 
We also observe that most of the errors from both methods are detection errors, this indicates that leveraging limited labels to improve the zero-shot pre-trained VL models and a better strategy to transform the detection results into action are potential directions to improve zero-shot methods. 

\begin{table}[t]
\centering
\caption{Comparison of exploration efficiency with and without commonsense reasoning on HM3D dataset. FrontierDist measures the average distance between chosen frontiers and the closest goal object. The other three metrics is the error rate of different error types.}
\vskip 0.1in
\label{tab:frontier dist}
\resizebox{\columnwidth}{!}{
    \begin{tabular}{lcccc}
    \toprule
         Model & \makecell[c]{FrontierDist \\ (meter)} & \makecell[c]{Exploration \\(\%)} & \makecell[c]{Detection \\(\%)} & \makecell[c]{Planning \\(\%)}\\
         \midrule
         GoW & 8.2   & 14.3 & \textbf{40.6}  & 12.1   \\
         ESC & \textbf{7.6}  & \textbf{10.6} & 40.8 & \textbf{9.5} \\
        \bottomrule 
    \end{tabular}
}
    \vspace{-15pt}
\end{table}

\begin{table}[t]
\centering
\caption{Comparison of results from ADMM and one-hot constraint for solving PSL program on MP3D dataset. We use object reasoning in the comparison.}
\vskip 0.1in
\label{tab:psl solver}
\resizebox{0.8\columnwidth}{!}{
    \begin{tabular}{lcccc}
    \toprule
         Solvers & SPL & SR & SoftSPL & Infer time\\
         \midrule
         ADMM & 12.9   & 27.0 & 22.1 & 2.13  \\
         One-hot & \textbf{13.7}  & \textbf{27.8} & \textbf{24.1} & \textbf{0.25} \\
        \bottomrule 
    \end{tabular}
    }
    \vspace{-12pt}
\end{table}

\textbf{Effect of different PSL solvers.} We compare the performance between ADMM and one-hot constraint for solving PSL optimization in Table~\ref{tab:psl solver}. We find that the results of the one-hot constraint are slightly better than ADMM, and it runs much faster for frontier selection (0.25 vs 2.13 seconds for each PSL inference). Since the final choice of the agent is one-hot, 
the target variables that optimize Eq.~\ref{eq: psl objective} in the one-hot space achieve the best satisfaction of the rules among all the possible choices of the agent.

\section{Related Work}

\noindent\textbf{Object Goal Navigation~}
Recently, there have been mainly two lines of work in object goal navigation. Most of the current SOTA methods use a pre-trained visual encoder~\cite{resnet,clip} to encode the egocentric images into feature vectors, then feed them into a navigation agent network trained by large-scale imitation learning or reinforcement learning~\cite{Ye_2021_ICCV,Treasure_Hunt,khandelwal2022:embodied-clip,Habitat-Web, procthor, chen2022learning}.
The second line of work is to explicitly construct a semantic map and train a navigation policy based on the constructed semantic map from the training dataset to infer goal object location ~\cite{chaplot2020learning,chaplot2020object, min2022film, zheng2022jarvis}. 
Compared with these methods, our visual understanding and navigation policy does not need data from a specific environment for training. Instead, we leverage the prompt-based text-image grounding model for scene understanding and commonsense knowledge in large language models to reason on both object and room levels in a zero-shot manner. 

To solve the problems of supervised methods on generalization to new objects and environments, four recent works aimed at zero-shot object goal navigation~\cite{gadre2022cow, majumdar2022zson, zero_experience, procthor}.
\citet{majumdar2022zson, zero_experience} both train an image-goal navigation agent at scale in target environments and map the object goal to image-goal embedding space, which may not be generalized well to new datasets. 
\citet{gadre2022cow} use GradCAM~\cite{gradcam} with CLIP to localize goal objects, and frontier-based exploration~\cite{fbe} for zero-shot object goal navigation. 
But it only uses CLIP to localize the goal object, and the exploration decision is not conditioned on the scene context. In contrast, our work leverages a grounded vision-and-language pre-trained model to recognize all common objects and rooms. Our exploration and navigation decisions are made conditioned on these scene contexts via a large language model in a zero-shot manner.

\vspace{0.2ex}
\noindent\textbf{Commonsense Reasoning in Embodied Agents~}
Commonsense reasoning is an essential ability for AI to perform human-level intelligence in various tasks and has been introduced into embodied AI tasks. \citet{chaplot2020learning, chen2022weaklysupervised, chaplot2020object, min2022film} used in-domain data to train a navigation policy on a semantic map to help find objects in an environment. 
External object relation knowledge was also used for object navigation in small environments~\cite{yang2019visual,slim} and embodied procedural planning~\cite{Lu2022NeuroSymbolicCL}. For household task completion, \citet{zheng2022jarvis} incorporate different levels of commonsense into task completion process, and \citet{tidee} leverage in-domain semantic prior for room rearrangement. 
In our work, we aim to transfer the commonsense knowledge in large language models into the object goal navigation task in a zero-shot manner by expressing the commonsense knowledge as first-order rules and encoding them into a declarative templating language---Probablistic Soft Logic to help better exploration.

\vspace{0.2ex}
\noindent\textbf{Large Pre-trained Models for Embodied Agents~}
Benefiting from large-scale pre-training, large pre-trained models have been shown to excel in vision and language tasks~\cite{clip,li2021grounded,khashabi2022unifiedqa}, and recently have been used for embodied AI tasks, including object navigation~\cite{khandelwal2022:embodied-clip, gadre2022cow, majumdar2022zson}, task planning~\cite{saycan2022arxiv,chen2022nlmapsaycan,huang2022inner,huang2022language,min2022film,hlsm}, and vision-and-language navigation~\cite{shen2021much,shah2022robotic}. 
Among these works, \citet{shen2021much,khandelwal2022:embodied-clip} utilize CLIP~\cite{clip} vision embedding to improve the performance of object goal navigation and vision-and-language navigation. \citet{majumdar2022zson} leverage CLIP to project image and text into one goal embedding space for zero-shot object navigation. 
For language models, \citet{min2022film,hlsm,sharma-etal-2022-skill,zheng2022jarvis} fine-tune a pre-trained language model on annotated (task, sub-tasks) pairs to teach the language model for sub-task planning. \citet{saycan2022arxiv, chen2022nlmapsaycan, huang2022inner, huang2022language} feed several examples as prompt to language models to decompose a high-level goal into executable steps for a robot. 
Our work, instead, focuses on efficiently finding a goal object in unseen environments. And we leverage the pre-trained location-related commonsense in large language models and object and room-level context detected by a pre-trained grounded vision-and-language model to guide exploration in a zero-shot manner. 
\label{sec:related}

\section{Conclusion and Future Work}
\label{sec:conclusion}
In this paper, we propose a zero-shot object navigation framework, ESC, that leverages the pre-trained knowledge of the language-image grounding model and large language model. We introduce commonsense into frontier-based exploration as a soft constraint via PSL. The experiment results illustrate the efficacy and generalizability of our methods from different perspectives.

Our work establishes new state-of-the-art and explores the direction of using pre-trained commonsense knowledge in LLMs for object navigation. Future work can try to acquire more commonsense from LLMs, like the spatial relation between rooms for object navigation, and acquire different knowledge from LLMs for other embodied AI tasks. What is more, ESC uses a fixed strategy to combine commonsense knowledge. Improving the fixed strategy or relaxing the zero-shot constraint to limited finetuning to learn a frontier selection strategy is also a potential direction. 

\section*{Acknowledgements}
We thank Xuehai He, Jialu Wang, and the anonymous reviewers for their valuable feedback on this work. 


\bibliography{example_paper}
\bibliographystyle{icml2023}

\newpage
\clearpage
\appendix

\section{Implementation Details}
\subsection{GLIP Implementation Details} \label{sec:glip appendix}

We use pre-trained GLIP-L~\cite{li2021grounded} for all our experiments. For both object detection and room detection, we use 0.61 as the threshold of GLIP. 
For object detection in MP3D and HM3D dataset, we define the common indoor objects $\mathcal{O}_c$ as all the goal objects in MP3D dataset, which has included the goal objects in the HM3D dataset. Therefore, the object detection prompt for both datasets will be \textit{`chair. table. picture. cabinet. cushion. sofa. bed. chest\_of\_drawers. plant. sink. toilet. stool. towel. tv\_monitor. shower. bathtub. counter. fireplace. gym\_equipment. seating. clothes.'}. 

For object detection in the RoboTHOR dataset, we utilize \textit{`Bed,
    Book,
    Bottle,
    Box,
    Knife,
    Candle,
    CD,
    CellPhone,
    Chair,
    Cup,
    Desk,
    Table,
    Drawer,
    Dresser,
    Lamp,
    Fork,
    Newspaper,
    Painting,
    Pencil,
    Pepper Shaker,
    Pillow,
    Plate,
    Pot,
    Salt Shaker,
    Shelf,
    Sofa,
    Statue,
    Tennis Racket,
    TV Stand,
    Watch'} as common indoor objects, which are selected from the furniture and object categories defined by RoboTHOR~\cite{RoboTHOR} dataset. These objects, combined with the 12 goal objects will constitute the object prompt on the RoboTHOR dataset. 
For room detection on three datasets, we define the room prompt as `bedroom. living room. bathroom. kitchen.  dining room. office room. gym. lounge. laundry room.'. 

\subsection{PSL detailed explanation} \label{sec:psl appendix}

During PSL inference, each ground rule creates one or more hinge-loss potentials defined over logical rules, which are relaxed using Łukasiewicz continuous valued logical semantics:
\begin{equation}
    \phi (Y,X) = [max(0,l(Y,X))]^{p}
\end{equation}
where $l$ is a linear penalty function defined by PSL. We explain how the potential function is calculated and help determine unobserved variables through the example rule $r$ in Eq.~\ref{eq: obj reason 1}: 
\begin{equation} \label{eq: obj reason 1 cp}
\begin{aligned}
    \mathrm{IsCooccur(Goal, Object)} \\
    \land \ \mathrm{IsNearObj(Frontier, Object)} \\ 
    \longrightarrow \mathrm{ChooseFrontier(Frontier)}
\end{aligned}
\end{equation}

This can be transformed into the following logic form:
\begin{equation}
\begin{aligned}
    \neg(\mathrm{IsCooccur(Goal, Object)} \\
    \land \ \mathrm{IsNearObj(Frontier, Object)} \\ 
    \lor \mathrm{ChooseFrontier(Frontier)}
\end{aligned}
\end{equation}

Given two grounded atoms $A_1, A_2 \in [0,1]$, the formulas for the Łukasiewicz relaxation of the logical conjunction ($\land$), disjunction ($\lor$), and negation ($\neg$) are as follows:
\begin{equation} \label{eq:potential cal}
\begin{aligned}
    A_1 \tilde{\land} A_2 &= \mathrm{max}\{0, A_1+A_2-1\}\\ 
    A_1 \tilde{\lor} A_2 &= \mathrm{min}\{A_1+A_2, 1\} \\ 
    \tilde{\neg} A_1 &= 1 - A_1 
\end{aligned}
\end{equation}

From Eq.~\ref{eq:potential cal}, by noting the ground atom as 
\begin{equation}
\begin{aligned}
    \mathrm{IsCooccur(Goal, Object)} = x_1 \\
    \mathrm{IsNearObj(Frontier, Object)} = x_2\\
    \mathrm{ChooseFrontier(Frontier)} = y_1 
\end{aligned}
\end{equation}
we can calculate the true value of a ground rule as :
\begin{equation}
\begin{aligned}
    \mathrm{min}\{1,(1-(x_1+x_2-1)) + y_1\} \\
    = \mathrm{min}\{1, 2-(x_1+x_2) + y_1\}
    \\
\end{aligned}
\end{equation}
The \textit{distance to satisfaction} of the rule is defined as 
\begin{equation}
\begin{aligned}
   & \ \ \ \phi (x_1,x_2,y_1) \\
    &=1 - \mathrm{min}\{1, 2-(x_1+x_2) + y_1\} \\
    &= \mathrm{max}\{0, x_1+x_2-y1-1\} 
\end{aligned}
\end{equation}
For instance, when $x_1=0.8$ and $x_2=0.8$, $\phi (x_1,x_2,y_1)=0$ only when $y_1>0.6$. And this will force the agent to choose a frontier when it's next to an object that the goal object is likely to be near. 




\subsection{LLM Details}  \label{sec:llm appendix}
For commonsense reasoning in object navigation, we choose the following two language models for zero-shot navigation inference. 

\noindent
\textbf{Deberta v3}~\cite{he2021debertav3} (which we will refer to as Deberta for simplicity hereafter) utilizes replaced token detection objective and disentangled attention mechanism for pre-training and achieved SOTA performance on a wide range of natural language understanding tasks. 
The Deberta v3 is first pre-trained on a commonsense reasoning QA (CSQA) dataset. During pre-training, each `question+candidate answer' pair in the dataset is fed into the Deberta model and gets the embedding $\mathbf{v}\in \mathcal{R}^{d}$ of $[\mathrm{CLS}]$ token from the output, which is then projected to a score $s_{i}$ with learned weights. 

During inference, we design two questions: `What is a $Goal Object$ likely to be near?' and `If you want to find a $Goal Object$, where should you go?' for object reasoning and room-level reasoning. 
The candidate rooms and objects in $\{o_g\} \bigcup \{o_c\}$ will be the candidate answers. The predicted scores $s_{i}$ will be linearly normalized into $\mathrm{0,1}$ and be the values of $\mathrm{IsCooccur}$ in the rules. 

\textbf{ChatGPT}~\cite{instrucGPT} is a recently release pre-trained conversational LLM. It is capable of answering questions in a reasonable way. For ChatGPT, we input the language prompt as `Among $P_{r}$ / $P_{o}$, can you give the scores of likelihood to find a $Goal Object$ inside / nearby?'. Here $P_{r}, P_{o}$ stands for room prompts and object prompts of GLIP respectively. The model will generate a series of scores that can be directly used as the values of $\mathrm{IsCooccur}$. 

\subsection{Navigation Algorithm and Deterministic Policy} \label{sec: nav alg appendix}
 \begin{algorithm}[t]
 \caption{Navigation algorithm}
 \label{alg:nav alg}
 \begin{algorithmic}
 \STATE \textbf{Input:} Goal $G$
 \STATE \textbf{Initialize:} frontier $F=None$, navigation map $M_{nav}$, semantic map $M_{sem}$, map update module $\mathrm{MAP}$, GoalDetected
 
 \STATE Object reasoning $\{R^{o}_{G,i}=\mathrm{LLM}(G, o_{i}, P^{LLM}_{o})\}$
 \STATE Room reasoning $\{R^{r}_{G,i}=\mathrm{LLM}(G, r_{i}, P^{LLM}_{r})\}$
 \STATE Look around and initialize $M_{sem}, M_{nav}$.
 \WHILE{not GoalDetected}
 \STATE $\{o_{t,i}, b^{o}_{t,i}\}=\mathrm{GLIP}(I_{t}, P_{o})$
   \STATE $\{r_{t,i}, b^{r}_{t,i}\}=\mathrm{GLIP}(I_{t}, P_{r})$
   \STATE $M_{sem} = \mathrm{MAP}(\{o_{t,i}, b^{o}_{t,i}\}, \{r_{t,i}, b^{r}_{t,i}\}, M_{sem})$
   \STATE $M_{nav} = \mathrm{MAP}(\{I_{t,i}\}, M_{nav})$
   \IF {$G$ in $\{o_{t,i}\}$}
   \STATE GoalDetected = True
   \STATE break
   \ENDIF
 \IF{reached $F$ or $F$ is $None$}
\STATE $F$ = $\mathrm{PSL}(\{R^{o}_{G,i}\}, \{R^{r}_{G,i}\}, M_{sem}, M_{nav})$
\ELSE 
 \STATE Navigate to $F$
 \ENDIF
 \ENDWHILE
 \IF{GoalDetected}
 \STATE Navigate to $G$
 \ENDIF
 \end{algorithmic}
\end{algorithm}
The navigation policy is illustrated in Alg.~\ref{alg:nav alg}. The agent will first perform object-level and room-level reasoning about the goal object. Then it will look around and initialize the semantic map and navigation map. During the navigation process, it will perform semantic scene understanding with GLIP at each step and update the information on the semantic/navigation map. It will select frontiers using PSL based on commonsense reasoning, semantic map, and navigation map. After the agent detect a goal object, it will directly navigate toward it. 

Under the general navigation policy, we also design several local policies to address the specific constraints encountered in different datasets and benchmarks. For MP3D and HM3D, the depth input is limited to 5 meters. Therefore, we design a long-distance goal policy to deal with the situation where a detected object is 5 meters away. If the object is a goal object, the agent will keep navigating in its direction until it's within 5 meters. If it's not a goal object, it will not be recorded in the semantic map. 

For RoboTHOR, there is no GPS input to the agent. Therefore, when the agent is facing a wall and takes a move-forward action, it will believe it moves 0.25 meters while it actually stays in the original place. To mitigate this, we calculate the difference of the depth input of the last step and the current step to judge if the agent is moving. 

\section{Dataset Details}  \label{sec:dataset appendix}
In MP3D~\cite{Matterport3D} dataset, there are 21 target objects: {\fontfamily{qcr}\selectfont
chair, table, picture, cabinet, cushion, sofa,
bed, chest\_of\_drawers, plant, sink, toilet, stool,
towel, tv\_monitor, shower, bathtub, counter, fireplace, gym equipment, seating, clothes}. In HM3D~\cite{ramakrishnan2021hm3d} dataset, there are 6 target objects: {\fontfamily{qcr}\selectfont
chair, sofa, plant, bed, toilet, and tv\_monitor}. In the RoboTHOR dataset, there are 12 target objects: {\fontfamily{qcr}\selectfont
AlarmClock, Apple, BaseballBat, BasketBall, Bowl, GarbageCan, HousePlant, Laptop, Mug, SprayBottle,
Television, Vase}. 


\section{More Results}  \label{sec:more results appendix}


\subsection{Results per category}

\begin{figure*}[h]
  \centering
     \begin{subfigure}
         \centering
         \includegraphics[width=0.2\textwidth]{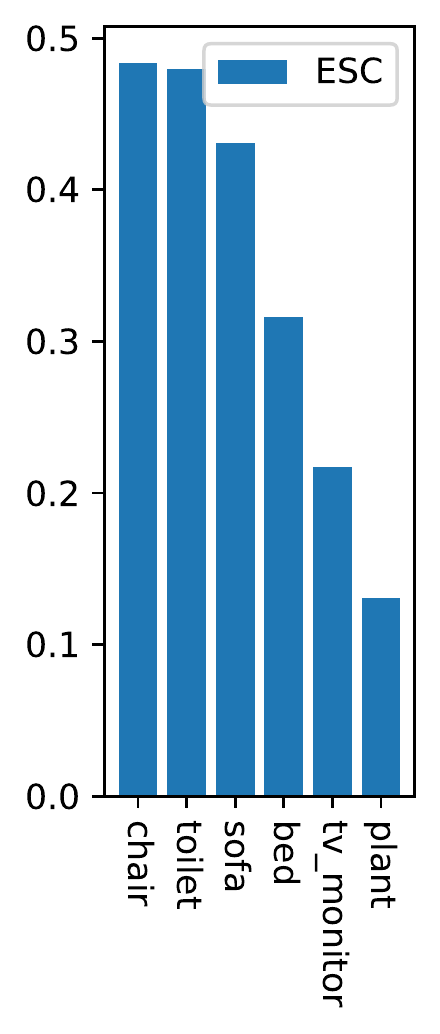}
     \end{subfigure}
     \begin{subfigure}
         \centering
         \includegraphics[width=0.49\textwidth]{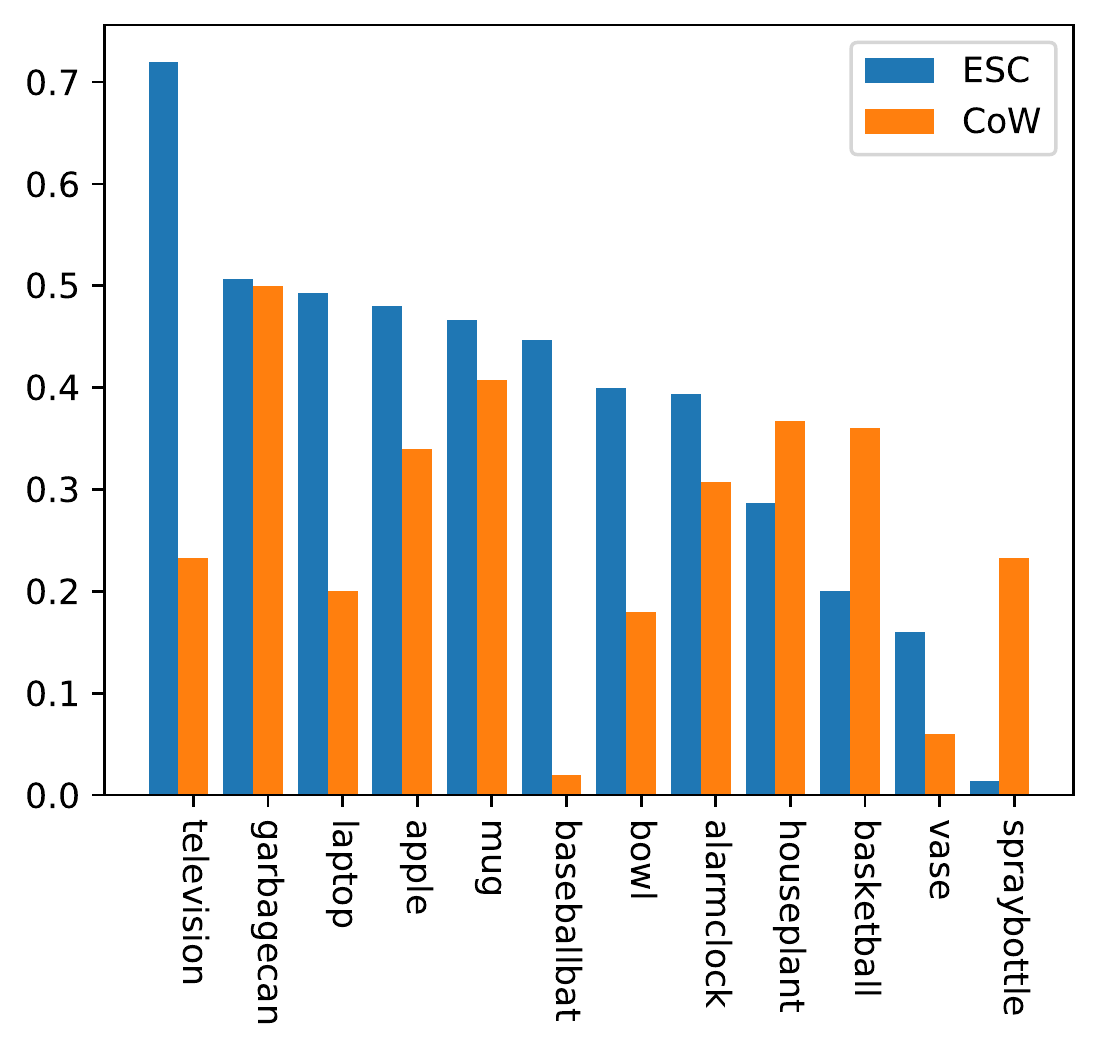}
     \end{subfigure}
  \vspace{-2ex}
  \caption{
A demonstration of the success rate of each goal category on HM3D (left) and RoboTHOR (right) datasets of ESC method and CoW.}
  \label{fig:result_cat}
\end{figure*}
We demonstrate the success rate of each goal category on HM3D and RoboTHOR datasets here and compare our ESC with CoW method in Fig.~\ref{fig:result_cat}.
From the results, we first observe that ESC performs well in all the categories in HM3D dataset. ESC performs better than CoW on most of the object goals on RoboTHOR, except houseplant, basketball and spray bottle. There are mainly two reasons. First, CLIP still has great object localization ability on certain objects. Second, the smaller exploration space of RoboTHOR requires less exploration ability, so our method has less advantage.

\begin{figure*}[h]
  \centering
         \includegraphics[width=\textwidth]{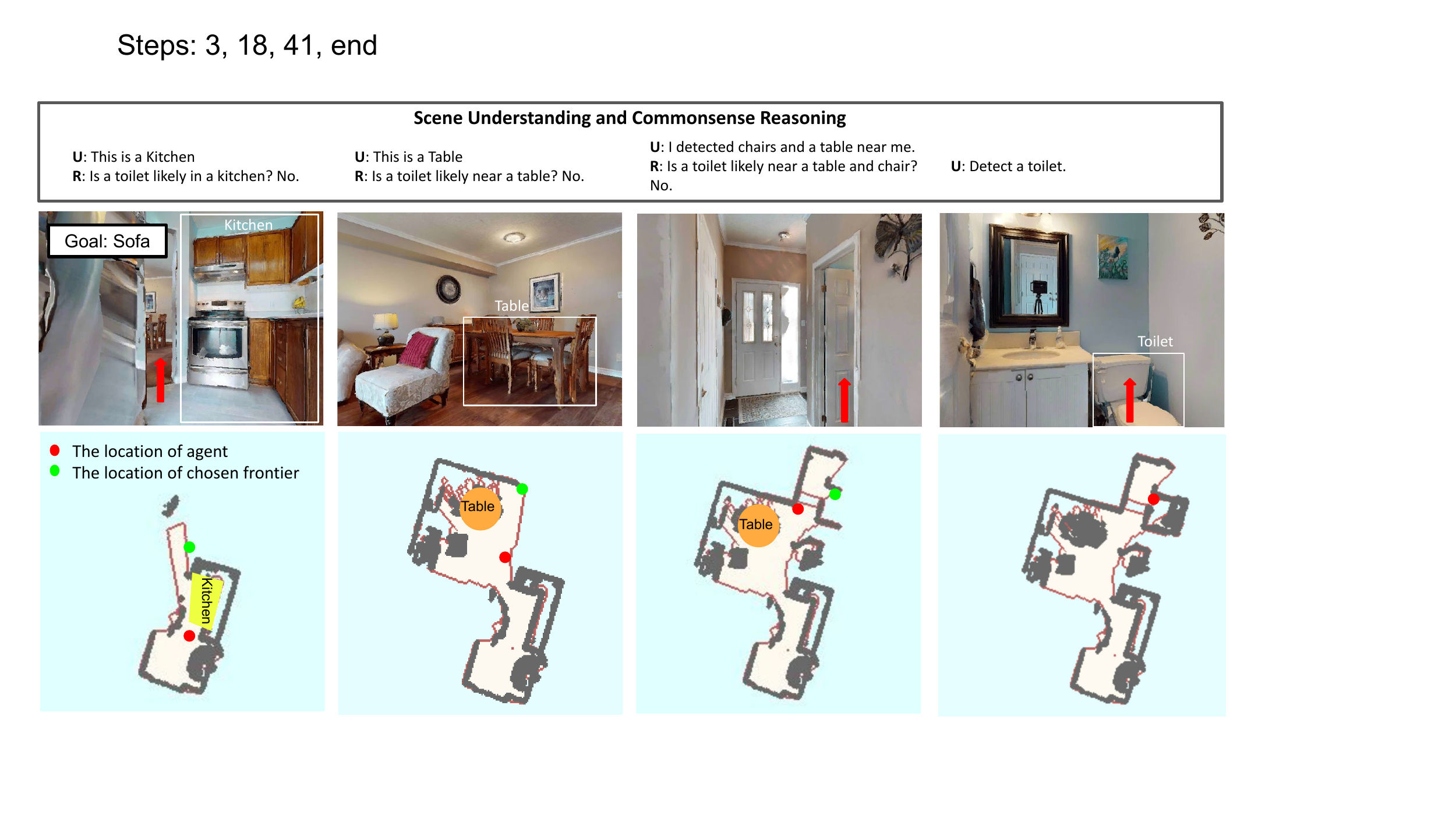}
  \caption{
  An example shows how commonsense reasoning helps the agent choose better frontiers that lead the agent to the goal `toilet'. `\textbf{U}' means scene understanding and `\textbf{R}' means commonsense reasoning. }
  \label{fig:case study 1}
\end{figure*}

\subsection{Case study}
To give a more intuitive view of how our model work in the object navigation process, we visualize an example in Fig.~\ref{fig:case study 1}. The agent chooses 3 frontiers during the navigation process as the green points show. First, the agent detects a kitchen and performs commonsense reasoning that the toilet is not likely in the kitchen. Therefore it selects a frontier with a certain distance from the kitchen. When it gets out of the kitchen, it detects a table and several chairs, which are also not likely to be near a toilet. So, the agent selects two frontiers far from them, which avoids useless exploration and helps it find the toilet.


\end{document}